# BRAIN AGE PREDICTION BASED ON RESTING-STATE FUNCTIONAL CONNECTIVITY PATTERNS USING CONVOLUTIONAL NEURAL NETWORKS

*Hongming Li[1], Theodore D. Satterthwaite[2], Yong Fan[1]*

[1]Department of Radiology, Perelman School of Medicine, University of Pennsylvania, Philadelphia, PA, 19104, USA
[2]Department of Psychiatry, Perelman School of Medicine, University of Pennsylvania, Philadelphia, PA, 19104, USA

## ABSTRACT

Brain age prediction based on neuroimaging data could help characterize both the typical brain development and neuropsychiatric disorders. Pattern recognition models built upon functional connectivity (FC) measures derived from resting state fMRI (rsfMRI) data have been successfully used to predict the brain age. However, most existing studies focus on coarse-grained FC measures between brain regions or intrinsic connectivity networks (ICNs), which may sacrifice fine-grained FC information of the rsfMRI data. Whole brain voxel-wise FC measures could provide fine-grained FC information of the brain and may improve the prediction performance. In this study, we develop a deep learning method to use convolutional neural networks (CNNs) to learn informative features from the fine-grained whole brain FC measures for the brain age prediction. Experimental results on a large dataset of resting-state fMRI demonstrate that the deep learning model with fine-grained FC measures could better predict the brain age.

## 1. INTRODUCTION

Brain age predicted based on biological phenotypes, such as anatomical and functional measures derived from neuroimaging data, and its deviation from the chronological age, could potentially serve as biomarkers for characterizing the typical brain development and clinical neuropsychiatric disorders [1, 2]. In the past years, computational neuroanatomy analytic tools and machine learning techniques have been adopted to predict the brain age based on structural/functional neuroimaging data [1-4]. Voxel-based morphometry (VBM) measures derived from structural MRI data have demonstrated promising performance for predicting the brain age in conjunction with different machine learning algorithms, such as linear support vector regression (SVR) [1, 2] and non-linear kernel based methods [5]. Cortical morphometry measures, such as cortical thickness and cortical surface area, have also been adopted for the brain age prediction in conjunction with the SVR [4]. Convolutional neural networks (CNNs) have recently been applied to raw structural MRI data for the age prediction and achieved comparable prediction performance as prediction models built upon the VBM measures [3].

Functional connectivity (FC) measures derived from resting-state fMRI (rsfMRI) data have advanced our understanding of the human brain functional organization and enable us to investigate both the typical brain development and neuropsychiatric disorders [6, 7]. Several studies have investigated the potential of FC measures derived from rsfMRI data for the brain age prediction [1, 8, 9]. In these studies, FC measures were derived from rsfMRI data based on regions of interest (ROIs) defined by different parcellation methods and the SVR was adopted to build age prediction models. For instance, FC measures between 160 ROIs in conjunction with the SVR were utilized to predict the brain age of subjects from age 7 to 30 years and demonstrated promising performance for characterizing the brain maturation [1]. Furthermore, several studies have also demonstrated that promising brain age prediction performance could be achieved by applying CNNs to brain networks [10].

These age prediction studies based on rsfMRI data have demonstrated the feasibility of rsfMRI data for the brain age prediction [1, 8, 9]. However, most of the existing studies focus on coarse-grained FC measures between different brain ROIs, which may sacrifice fine-grained FC information across the whole brain. On the other hand, data-driven brain decomposition methods have provided alternative tools for charactering intrinsic connectivity networks (ICNs) and estimate functional network connectivity and are favored for modeling many-to-many mapping between brain regions and functions [11]. Although ICNs have been widely adopted in FC studies, its potential to aid the age prediction remains largely unknown.

In this study, voxel-wise FC measures between voxels and intrinsic functional networks (ICNs) across the whole brain are utilized for the brain-age prediction. These dense FC maps could encode fine-grained FC information across the whole brain, and might improve the prediction performance. To learn fine-grained FC measures that are informative for the age prediction, we adopt deep CNNs to learn hierarchical FC patterns from the dense FC maps of all ICNs. Particularly, dense FC maps of multiple ICNs are used as multi-channel input to the deep CNNs, and the convolutional filters across multiple FC maps are optimized to learn hierarchical FC patterns for the brain age prediction using deep residual networks [12]. We have validated the proposed deep learning method for the brain age prediction by applying it to a large rsfMRI dataset obtained from the Philadelphia Neurodevelopmental Cohort (PNC) [13]. The experimental results have demonstrated that our deep learning model could obtain promising brain age prediction performance.

## 2. METHODS

To incorporate fine-grained FC information into the brain age prediction model, we first identify subject-specific ICNs and compute the whole brain voxel-wise FC measures for each ICN. Then, each subject's whole brain FC measures of all ICNs are stacked as multiple channels to form a 4D image as input to our deep CNNs model.

### 2.1. Voxel-wise whole brain FC measures

To calculate the voxel-wise whole brain FC measures, we first identify subject-specific, nonnegative, sparse ICNs based on rsfMRI data using a collaborative non-negative matrix factorization based algorithm [14, 15]. This method could obtain subject-specific, non-negative ICNs without losing inter-subject correspondence. In this study, we computed 56 ICNs for each subject, and the number of ICNs was estimated automatically by MELODIC of FSL based on the Laplace approximation criterion. Example ICNs and 3D rendering of all the 56 ICNs at a group level are illustrated in Fig. 1.

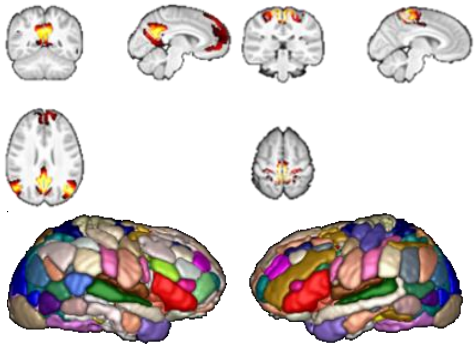

Fig. 1. Example ICNs obtained for calculating voxel-wise whole brain FC measures. Example ICNs (top row), and 3D rendering of all ICNs encoded in different colors (bottom row).

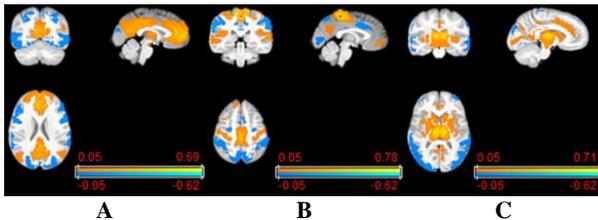

Fig. 2. Example voxel-wise whole brain FC measures of ICNs: (A) default mode network, (B) somatomotor network, and (C) cingulo-opercular network.

Based on the subject-specific ICNs, the whole brain voxel-wise FC measures for each ICN are calculated as Pearson correlation coefficients between its corresponding time course and voxel-wise rsfMRI signals across the whole brain, and transferred to Fisher's z-scores so that both local and long range FC measures are obtained for each ICN. Example voxel-wise FC measures of different ICNs are illustrated in Fig. 2.

### 2.2. Brain age prediction using deep CNNs

Given voxel-wise whole brain FC measures of all the ICNs of each subject, they are stacked as a multi-channel 4D FC image. Each subject's 4D FC image is used as input to the deep CNNs for building a brain age prediction model. As the whole brain FC measures preserve the spatial information as the original fMRI image, the convolutional filters could extract informative features that encode the interaction of FC measures across all the ICNs for each voxel, and the deep structure of the network facilitates a hierarchical high-level feature extraction. Both the feature extraction and prediction are optimized simultaneously for the brain age prediction. The architecture of our deep CNNs is illustrated in Fig. 3.

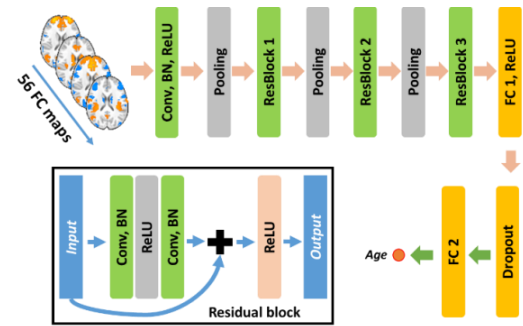

Fig. 3. Schematic diagram of deep CNNs for FC patterns based brain age prediction.

As illustrated in Fig. 3, our deep model contains 1 convolutional layers (Conv), followed by 3 residual blocks (ResBlock), 1 fully connected layers (FC1), and an output layer (FC2) for brain age prediction, in addition to pooling and dropout layers. The residual network structure has been adopted widely since its invention [12] and achieved promising performance in many challenging pattern recognition tasks. The residual connection would also accelerate the convergence and improve the performance of the CNNs. Rectified linear units (ReLU) is used as the nonlinear activation function for the convolutional and fully connected layers, batch normalization (BN) is adopted to accelerate deep network training, and max pooling layers are adopted to obtain features at multiple scales. Euclidean loss between the predicted age and the chronological age is used to optimize the whole network.

### 2.3. Visualization of the deep CNNs

To understand the deep learning models, we carry out a sensitivity analysis to determine how changes in FC maps of ICNs impact the deep learning model with respect to the age prediction based on *N* testing subjects using a principal component analysis (PCA) based sensitivity analysis method [16]. Particularly, with the trained deep CNNs fixed, FC maps of *n* ICNs are excluded (i.e., its values are set to zero) one by one from the input to the deep learning model and changes in the predicted ages are recorded. Once all the changes in the brain age prediction with respect to all ICN are obtained for all *N* testing subjects, we obtain a change matrix of *n* x *N*, encapsulating changes of the brain age prediction. We then apply PCA to the change matrix to identify principle components (PCs) that encode main directions of the age prediction changes with respect to FC maps of ICN.

We also use t-Distributed Stochastic Neighbor Embedding (t-SNE) to reduce dimensionality of high dimensional output of the last fully connected layer of the deep learning models [17]. The high dimensional outputs of different subjects are projected onto a 2D plane to visually inspect spatial distribution of the learned features at the last layers of the deep learning model.

## 3. RESULTS

### 3.1. rsfMRI dataset

A dataset consisting of rsfMRI scans of 983 subjects from the PNC dataset [13] (ages from 8 to 22) was used to evaluate the performance of the proposed method. The fMRI data were preprocessed using an optimized procedure, including slice timing, confound regression, and band-pass filtering [18]. MELODIC of FSL was used to automatically estimate the number of ICNs with the Laplace approximation criteria, and the estimated number of ICNs was 56. Finally, for each subject 56 subject-specific non-negative ICNs were identified to compute the whole brain FC measures for the brain age prediction [14]. To reduce the computation burden, the whole brain FC measures were down-sampled at a spatial resolution of 4x4x4mm$^3$.

### 3.2. Experimental setting

Our deep learning model's network architecture is illustrated in Fig. 3, with 1 Conv layer, 3 ResBlocks, 1 FC layer, and an output layer. In particular, the Conv layer contained 64 kernels, while the ResBlock 1, 2, and 3 contains 64, 128, and 128 kernels respectively. The kernel size for all the kernels was 3×3×3. A stride of 2 and kernel size of 2 was used for the max pooling layer. The fully connected layer FC1 contained 256 nodes, the 256-dimensional feature vector was fed to output layer FC2 with 1 output node for age prediction. A dropout operation with a ratio of 0.5 was applied before the features were fed into the last FC layer.

The deep learning model was optimized using stochastic gradient descent (SGD) algorithm, the momentum was set to 0.9, and the base learning rate was set to $1 \times 10^{-4}$. The learning rate was updated using a stepwise policy by dropping the learning rate by a factor of 0.1 after every 10000 steps. The maximum iteration of the training procedure was set to 30000. The batch size was set to 16. The deep learning model was implemented using Caffe [19], and trained on a Nvidia Titan X (Pascal) graphics processing unit (GPU).

We compared the proposed deep learning model with sparse regularized least-squares regression (lasso) [20] models based on both coarse-grained inter-ICN FC measures and whole brain FC measures. Particularly, the inter-ICN FC measures of each subject contained 1540 elements between pairs of ICNs. For the whole brain FC measures, all the 56 FC maps of each subject were flattened and concatenated as a vector of features (with 495600 elements), and the feature dimensionality was reduced to ~80000 by removing features with relatively smaller Pearson correlation coefficients with the chronological age in training datasets (*p*>0.05). Parameters of the lasso models were optimized by nested 5-fold cross-validation. For the inter-ICN FC measures, we also used the BrainNetCNN, particularly the E2Enet_sml, to build a brain age prediction model [10]. All these models were evaluated under the same 5-fold cross-validation setting. Pearson correlation coefficient and mean absolute error (MAE) between the predicted brain age and the chronological age were used to evaluate the performance of different models.

Table1. Prediction performance of different age prediction models.

| Models | Features | Correlation | MAE (years) |
|---|---|---|---|
| lasso | inter-ICN FC | 0.417±0.053 | 2.45±1.78 |
| E2Enet-sml | inter-ICN FC | 0.361±0.036 | 3.02±2.28 |
| lasso | whole brain FC | 0.530 ±0.064 | 2.32±1.61 |
| CNNs | whole brain FC | 0.614±0.059 | 2.15±1.54 |

### 3.3. Experimental results

Quantitative evaluation results of all these models are summarized in Table 1. These results demonstrated that the prediction models built upon the whole brain FC measures outperformed those built upon the inter-ICN FC measures with respect to both the correlation and MAE measures, indicating that the fine-grained whole brain FC measures were more informative for the brain age prediction. The proposed deep learning model worked better than the lasso model built upon the whole brain FC measures, indicating that the hierarchical features learned by the deep learning model better characterized the brain developmental information than the original FC measures.

As shown in Fig. 4, the sensitive analysis revealed ICNs whose changes were more sensitive than others to the age prediction. Particularly, top 5 ICNs with the largest magnitudes in the first PC are shown in Fig. 4. Three of them were corresponding to the cingulo-opercular network, the default model network, and the attention network whose FC measures change along with the age [21].

The t-SNE projection results as shown in Fig. 5 further demonstrated that the features learned by our method from dense FC measures contained information more consistent with the age distribution than other features, indicating that our method could improve the prediction performance.

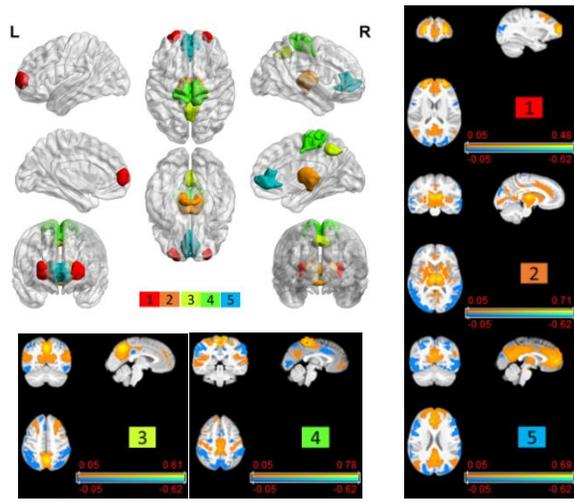

Fig. 4. Top 5 ICNs (top left) that were more sensitive to the prediction model and their corresponding whole brain FC measures (top right and bottom row).

## 4. CONCLUSION

In this study, we proposed a deep learning method to learn informative features from the whole brain voxel-wise FC measures using convolutional neural networks for the brain age prediction. Our study demonstrated that the fine-grained FC information in conjunction with the deep CNNs could improve the brain age prediction. Our deep learning model is flexible to integrate multimodal information. We expect that the age prediction performance could be further improved if multi-modality information is used to build the age prediction model.

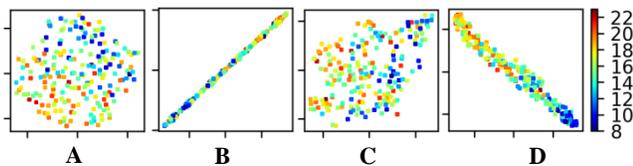

Fig. 5. t-SNE visualization of (A) the inter-ICN FC measures, (B) features learned by E2Enet-sml, (C) original whole brain FC measures, and (D) features learned by CNNs. Colors of points refer to their chronological age shown in the colorbar.

## ACKNOWLEDGEMENTS

This work was supported in part by National Institutes of Health grants [EB022573, CA223358, MH107703, DK114786, DA039215, and DA039002].